\pdfoutput=1
\documentclass[11pt]{article}

\usepackage{arabtex}
\usepackage{utf8}
\setcode{utf8}

\usepackage{multirow}

\usepackage{graphicx}
\usepackage{subcaption}
\usepackage{amsmath}

\usepackage{tablefootnote}

\usepackage{EMNLP2023}

\usepackage{times}
\usepackage{latexsym}

\usepackage[T1]{fontenc}

\usepackage[utf8]{inputenc}

\usepackage{microtype}

\usepackage{inconsolata}

\title{CATT: Character-based Arabic Tashkeel Transformer}

\author{Faris Alasmary \\
  Abjad Ltd. \\
  \texttt{falasmary@abjad.com.sa} \\\And
  Orjuwan Zaafarani \\
  Abjad Ltd. \\
  \texttt{oturki@abjad.com.sa} \\\And
  Ahmad Ghannam \\
  Abjad Ltd. \\
  \texttt{aghannam@abjad.com.sa} \\}

\begin{document}
\maketitle
\begin{abstract}

Tashkeel, or Arabic Text Diacritization (ATD), greatly enhances the comprehension of Arabic text by removing ambiguity and minimizing the risk of misinterpretations caused by its absence.
It plays a crucial role in improving Arabic text processing, particularly in applications such as text-to-speech and machine translation.
This paper introduces a new approach to training ATD models.
First, we finetuned two transformers, encoder-only and encoder-decoder, that were initialized from a pretrained character-based BERT.
Then, we applied the Noisy-Student approach to boost the performance of the best model.
We evaluated our models alongside 11 commercial and open-source models using two manually labeled benchmark datasets: WikiNews and our CATT dataset.
Our findings show that our top model surpasses all evaluated models by relative Diacritic Error Rates (DERs) of 30.83\% and 35.21\% on WikiNews and CATT, respectively, achieving state-of-the-art in ATD.
In addition, we show that our model outperforms GPT-4-turbo on CATT dataset by a relative DER of 9.36\%.
We open-source our CATT models and benchmark dataset for the research community\footnote{https://github.com/abjadai/catt}.






\end{abstract}
\section{Introduction}

The Arabic language is characterized by its rich morphology and complex syntactic structure.
One of the unique features of Arabic is the use of diacritics or Tashkeel, which are small marks above or below the letters that indicate vowels or other phonetic aspects of pronunciation.
These diacritics are favorable for understanding the meaning of words, as their absence can lead to ambiguities and misinterpretations.
They are also crucial in improving performance of applications such as text-to-speech and machine translation \cite{fadel-nov2019}.
The diacritics can be affected by the context of the sentence as shown in the following example:

\begin{center}
1. \<سَاقَ الرَّجُلُ السَّيَّارَةَ>

    \begingroup
        \fontsize{6}{8}\
            Translation: The man drove the car. 
    \endgroup

2. \<سَاقُ الرَّجُلِ بِهَا جُرُوحٌ>

    \begingroup
        \fontsize{6}{8}\
            Translation: The man's leg has wounds.
    \endgroup
\end{center}
In the first sentence, "\<ساق>" or "Saqa" is interpreted as "drove", denoting an action. However, when the context changes, the same word, now pronounced as "Saqu" takes on a completely different meaning, becoming a noun that translates to "leg".
This change highlights the crucial role of diacritics in quickly clarifying the meanings of sentences, as the characters remain the same.
At the same time, the pronunciation varies depending on the context.
In the previous example, the meaning of the word "\<ساق>" can be comprehended even without diacritics, as long as the reader considers the complete sentence and understands the context provided by the surrounding words.
This fact raises the following question: "Will a pretrained BERT model help in improving the ATD models?".

In this paper, we propose a training strategy based on a pretrained character-based BERT \cite{kenton2019bert,liu2019roberta}, and a self-training approach called Noisy-Student (NS) \cite{xie2020self}.
Throughout the paper, we will answer the following research questions:
\begin{itemize}
  \item \textbf{RQ1:} Does the ATD model benefit from Masked Language Model (MLM) pretraining?
  \item \textbf{RQ2:} Does training ATD model for more iterations help?
  \item \textbf{RQ3:} Is the NS approach effective in ATD models?

\end{itemize}

\section{Related Work}

Previous research has investigated a broad spectrum of approaches to address the diacritization task, beginning with rule-based methods, moving to classical machine learning models, and reaching sophisticated deep learning architectures \citep{almanea2021automatic}.
In addition, comprehensive experiments show that deep learning methods outperform non-neural techniques, particularly when substantial training data is available \citep{fadel-apr2019}.

\citet{fadel-nov2019} tested a refined version of the Tashkeela dataset \citep{zerrouki2017tashkeela,fadel-apr2019} using the Shakkala\footnote{https://github.com/Barqawiz/Shakkala} model.
They also trained a character-level RNN with a Block-Normalized Gradient (BNG) module.
The BNG technique normalizes gradients within each batch, potentially speeding up training and improving generalization \citep{bng2017yu}.

\citeposs{abbad2020multicomponents} ATD approach consisted of a three-part pipeline: a multi-layer LSTM and dense layers, a character-level rule-based corrector for specific error correction, and a word-level statistical corrector that leveraged context and distance information to resolve diacritization issues.
Furthermore, they developed an enhanced version of the system and named it Multilevel Diacritizer
\citep{abbad2021simple}.

\citet{madhfar2020effective} implemented 3 different ATD models.
The first one was a baseline model consisting of 3 deep Bidirectional Long Short-Term Memory (BiLSTM) layers.
The second model was an encoder-decoder with 3 LSTM layers for the encoder and 2 LSTM layers for the decoder.
The last model was based on Tacotron encoder \citep{wang2017tacotron} that uses CBHG module \citep{lee2017fully}.

\citet{alkhamissi-etal-2020-dd} proposed two architectures: the Two-Level Diacritizer (D2) and the Two-Level Diacritizer with Decoder (D3).
D3 builds upon the capabilities of D2 by accepting partially diacritized text as input.
These models have a word-level encoder as well as a character-level encoder.
The results of both encoders are combined by an attention mechanism and fed to a unidirectional LSTM layer to predict diacritics.

\citet{darwish2021arabic} created two Deep Neural Networks (DNNs); the first utilizes a character-based BiLSTM model with unique features for each character, while the second uses a word-level BiLSTM layer and a subsequent dense layer with Softmax activation.

\citet{Karim2021} studied the effect of varying the training dataset size.
Each time, they trained a BiLSTM model and evaluated its performance. The results demonstrated that error rates improve as the size of the training corpus increases.

\citet{alrfooh2023finetashkeel} finetuned a token-free multilingual model called ByT5 \citep{xue2022byt5} to perform Arabic text diacritization as a sequence-to-sequence task, similar to the translation task.

Recently, \citet{alike-method} introduced the Pre-FineTuned Token Classification for Arabic Diacritization (PTCAD) model.
This approach treats Arabic text diacritization as a downstream task for a pretrained BERT-like model.
The approach starts with a pretraining phase on linguistically relevant tasks, such as Part-of-Speech (POS) tagging and Segmentation, which are framed as Masked Language Modelling (MLM) tasks. This pretraining helps enhance the model's contextual understanding.
Then, it moves into a finetuning phase where diacritization is handled as a token classification task.
This phase leverages the contextual insights gained earlier to enhance diacritization accuracy.

Unlike \citeposs{alike-method} method, we consider a simpler approach where we directly pretrain a character-level BERT model with no further modifications or extra labeling.
\section{Dataset Preparation}
\subsection{Training Data}
\label{sec:training_data}

As shown by \citet{Karim2021}, training on larger dataset improves the performance of the ATD model.
We used the whole Tashkeela dataset \citep{zerrouki2017tashkeela} for training which consists of 1,658,325 samples.
Initially, we filtered out samples that had fewer than 6 characters or more than 1024 characters, considering both letters and diacritics as characters.
Next, we removed samples with a Diacritics-to-Letters (DTL) ratio of less than 60\%. We defined this ratio as follows:\\
\[
    DTL\; ratio\; = \frac{\#\; of\; diacritics}{\#\; of\; letters}
\]
In addition, we performed a cleaning process on each sentence in the filtered list, removing non-Arabic characters. This includes special characters, English letters, Arabic and Indian numerals, as well as punctuation marks in both English and Arabic.
After this cleaning process, the total number of remaining samples was 1,330,539.

To pretrain the character-based BERT, we scraped 18,543,025 data samples from various sources, including X and online news websites.
Training on this data will help the model to understand the Modern Standard Arabic (MSA) as well as the colloquial dialects.
To align with the architectural requirements of the ATD models for subsequent finetuning, we capped the maximum sequence length of the model at 1024 characters.
However, we set the maximum length of the training sentences during MLM pretraining to 512.
Consequently, all samples in the pretraining data were truncated at the last space character when the length of the sample exceeds 512 to preserve the context of the last word in the sample.

\subsection{Benchmark Data}
The Tashkeela \citep{zerrouki2017tashkeela} dataset contains data from different sources, including both MSA and classical Arabic.
Around 98.85\% of the Tashkeela dataset consists of content obtained from 97 books found in the Shamila\footnote{https://shamela.ws} library.
The Shamila library is an Islamic electronic library with hundreds of works covering Hadith, Fiqh, history, preaching, Islamic rules, and Arabic language \citep{zerrouki2017tashkeela}.
It can be misleading to assess ATD models using a portion of this dataset for the following reasons:
\begin{enumerate}
    \item Most of the dataset's books contain partial or complete citations from the Holy Quran and Hadith as well as from each other, which might lead to data contamination eventually impacting the evaluation results.
    
    \item Most resources in the dataset are written in classical Arabic.
    However, when evaluating ATD models for today's applications such as text-to-speech or machine translation, relying only on this dataset may lead to unreliable results.
    This is because the target users of these applications typically use MSA or colloquial dialects, which differ from classical Arabic.
\end{enumerate}
As a result, we created the CATT benchmark dataset.
This dataset comprises 742 sentences, which we scraped from an internet news source in 2023.
It covers multiple topics including science and technology, economics, politics, sports, arts, and culture.
The CATT dataset was manually diacritized by two expert native Arabic speakers and then validated by a third expert.
This dataset contains names of people and places in both Arabic and English.
As for the English names, they are written in Arabic letters and diacritized based on their pronunciation.
Also, the numbers in the sentences are written in textual form rather than the numeric form.
This helps in evaluating the models without the need for a text normalizer (TN).

Moreover, we used WikiNews \citep{darwish2017arabic} benchmark dataset to evaluate all models.
This dataset comprises 400 manually diacritized MSA sentences.
It covers multiple topics, most of which are similar to CATT's topics, from the years 2013 and 2014.

\begin{table}
    \centering
    \small
    \begin{tabular}{c|c}
        \hline
        \textbf{Class Name} & \textbf{Diacritic} \\
        \hline
        Fatha & \<بَ> \\
        Kasra & \<بِ> \\
        Dhamma & \<بُ> \\
        Tanween Fath & \<بً> \\
        Tanween Kasr & \<بٍ> \\
        Tanween Dhamm & \<بٌ> \\
        Shadda  & \<بّ> \\
        Shadda + Fatha & \<بَّ> \\
        Shadda + Kasra & \<بِّ> \\
        Shadda + Dhamma & \<بُّ> \\
        Shadda + Tanween Fath & \<بًّ> \\
        Shadda + Tanween Kasr & \<بٍّ> \\
        Shadda + Tanween Dhamm & \<بٌّ> \\
        Sukoon & \<بْ> \\
        No Tashkeel & <NT> \\
    \end{tabular}
    \caption{Arabic Diacritics}
    \label{table:tashkeel_classes}
\end{table}

\begin{table}[ht!]
    \centering
    \small
    \begin{tabular}{c|ccc}
        \hline
        \textbf{Data} & \textbf{Chars} & \textbf{Words} & \textbf{Lines}\\
        \hline
        BERT Pretraining & 2.06B & 359.96M & 18.54M \\
        Tashkeela & 213.86M & 42.43M & 1.33M \\
    \end{tabular}
    \caption{Data Summary (After Preparation)}
    \label{tab:data-summary}
\end{table}

\section{Experiments}
\begin{figure*}[ht!]
    \centering
    \includegraphics[width=0.8\linewidth]{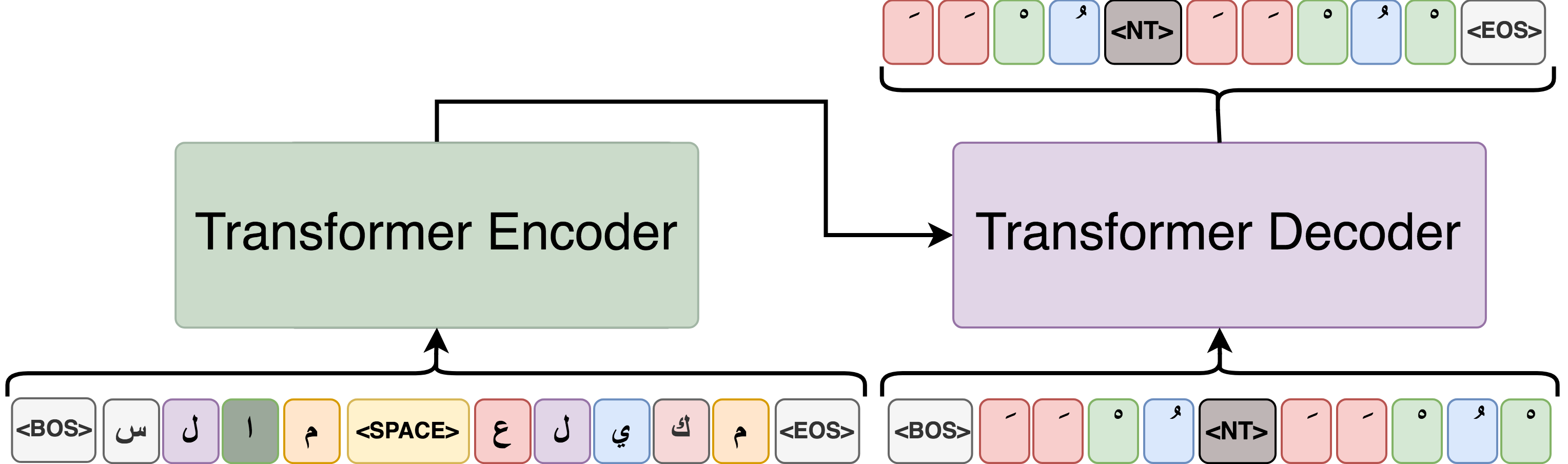}
    \caption{Encoder-Decoder (ED) Model}
    \label{fig:catt_encoder_decoder}
\end{figure*}
\begin{figure}
    \centering
    \includegraphics[width=0.7\linewidth]{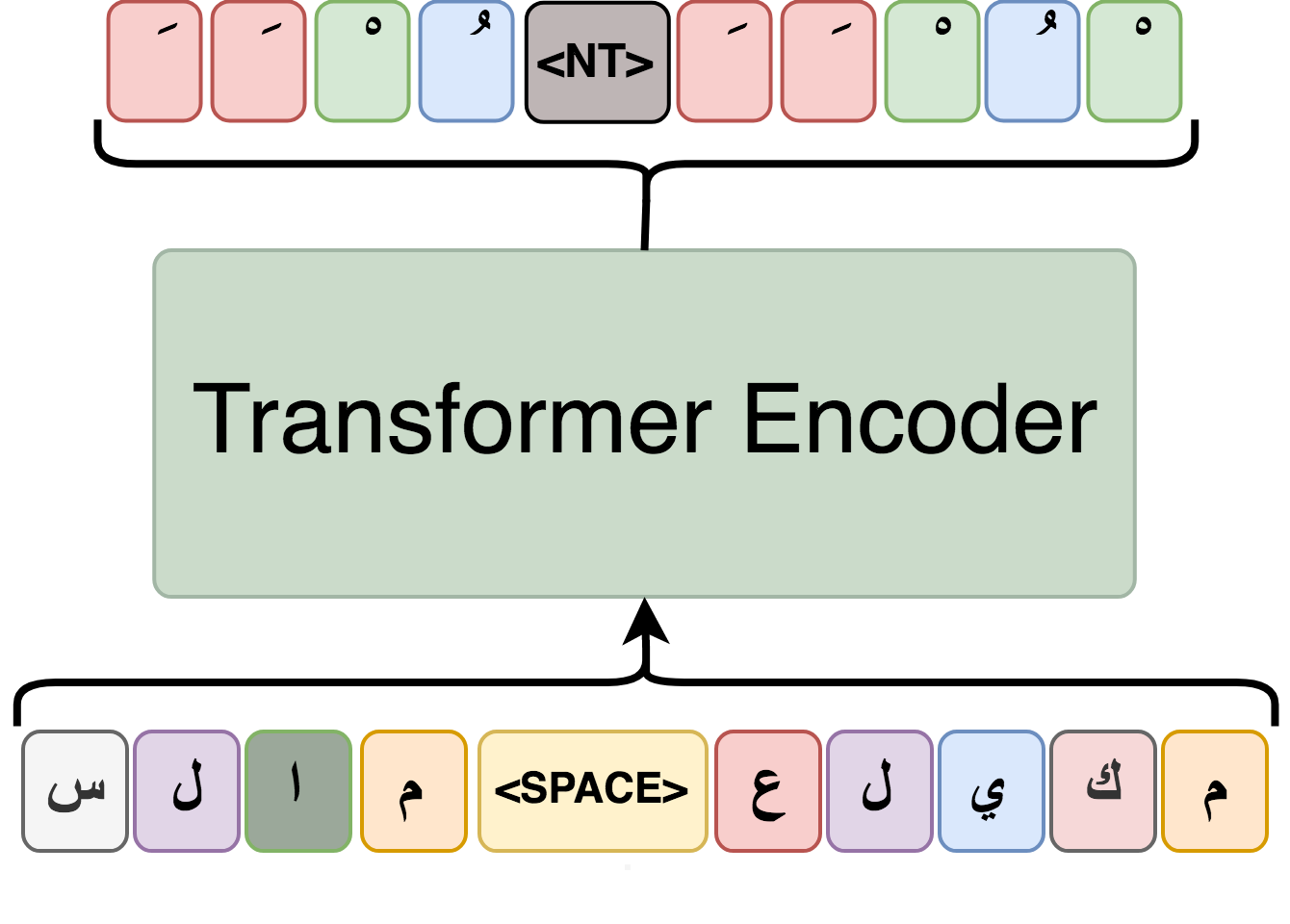}
    \caption{Encoder-Only (EO) Model}
    \label{fig:catt_encoder_only}
\end{figure}
We pretrained a character-based BERT model to find the effect of MLM pretraining on the diacritization performance.
The model has 6 layers with $d_{model} = 512$ and number of heads$\; = 16$.
The model was trained using MLM loss for 6 epochs with a batch size of 512, using the data shown in Table \ref{tab:data-summary}.

For ATD models, we selected two transformer architectures: Encoder-Decoder (ED) with 3 layers and Encoder-Only (EO) with 6 layers.
We set the number of layers to 3 in the ED model to ensure comparability with the EO model in terms of the total number of layers.
All our ATD models were trained with the following configurations: $d_{model} = 512$, number of heads$\; = 16$, batch size$\; = 32$, and dropout$\; = 10\%$.
We used the AdamW optimizer \cite{loshchilov2018decoupled} with a learning rate of $3 \times 10^-5$ and a weight decay of $1 \times 10^-2$.
Each model was trained on a single dedicated A100 GPU.
All ATD models in our experiments were trained for a maximum of 200 epochs with an early stopping criteria.

Generally, we define the input text \(X\) with a total number of characters \(T\) as a series of characters \(x_1, x_2, x_3, \ldots, x_T\) where each character represents an undiacritized Arabic letter.
Correspondingly, the output sequence \(Y\) consists of diacritics \(y_1, y_2, y_3, \ldots, y_T\) with each diacritic \(y_i\) associated with the respective letter \(x_i\).
In EO models, we express the relationship between letters and diacritics as follows:
\[P(y_i \mid  x_1, \ldots, x_T)\]
In other words, we predict the diacritic \(y_i\) conditioned only on the input text \(x_1, \ldots, x_T\).
In ED models, on the other hand, we consider the ATD task as a translation task, where the input text represents the source language and the output diacritics sequence represents the target language.
We express the relationship between letters and diacritics in ED models as follows:
\[P(y_i \mid  x_1, \ldots, x_T, y_1, \ldots, y_{i-1})\]
where the diacritic \(y_i\) is conditioned on both the input text \(x_1, \ldots, x_T\) and the previous diacritics \(y_1, \ldots, y_{i-1}\).
In fact, native Arabic speakers rely on both the textual content and diacritics to better disambiguate the intended meaning of the sentence.
For example, the sentence \<اشتريت لعبة> is a complete and valid Arabic sentence that can be interpreted as "I bought a toy" or "A toy was bought".
The only way to differentiate between them in the textual form is by adding diacritics as follows:
\<اشْتَرَيْتُ لُعْبَةً>
which means "I bought a toy"
or 
\<اشْتُرِيَتْ لُعْبَةٌ>
which means "A toy was bought".
In both cases, the diacritics of the second word heavily depend on the diacritics of the first word.
Therefore, by conditioning on both the input text and the previous diacritics, the model can achieve better performance.
Figure \ref{fig:catt_encoder_decoder} and Figure \ref{fig:catt_encoder_only} show both transformer architectures.

Our experiments involved training a total of four models using both the ED and EO architectures.
For both architectures, we used the pretrained character-based BERT as the basis for initializing the weights. Consequently, we obtained one ED model and one EO model with weights that reflect the knowledge encoded in the pretrained BERT.
Additionally, we trained the other two models, consisting of one ED model and one EO model, where the weights were randomly initialized.
These models allowed us to explore the impact of different weight initialization strategies on the performance and behavior of the models.

For each model in our experiments, we selected two checkpoints.
The first checkpoint was chosen after training the model for 5 epochs, while the second checkpoint was chosen as the best checkpoint achieved after training for a longer period.
The best checkpoint of ED model was at epoch 175 while the best checkpoint of EO was at epoch 192.
The purpose of this selection process was to study the impact of extended training duration on the models, even when they were exposed to the same amount of data. 

Moreover, we randomly sampled 1M sentences from the pretraining dataset to be pseudo-labeled using the NS \cite{xie2020self} technique.
We used the best checkpoints of both ED and EO models to pseudo-label two copies of the sampled data.
Finally, a new ED model as well as a new EO model were trained on Tashkeela data combined with the pseudo-labeled 1M sentences.
Both models' parameters were initialized from the best checkpoints. Table \ref{tab:noisy_student_data} shows the details of the combined dataset after the filtration process described in section \ref{sec:training_data}.
\begin{table}[ht!]
    \centering
    \small
    \begin{tabular}{c|ccc}
        \hline
        \textbf{Data} & \textbf{Chars} & \textbf{Words} & \textbf{Lines}\\
        \hline
        Tashkeela & 213.86M & 42.43M & 1.33M \\
        Tashkeela + NS (ED) & 292.01M & 56.21M & 2.22M \\
        Tashkeela + NS (EO) & 290.59M & 55.95M & 2.20M\\ 
    \end{tabular}
    \caption{The combined datasets using NS pseudo-labeling by both ED and EO models (After Preparation).}
    \label{tab:noisy_student_data}
\end{table}

\section{Results}

There are two methods to evaluate ATD models: one with Case Ending (CE) and one without Case Ending (No CE).
In the No CE approach, the diacritic on the last letter is excluded during performance evaluation, while the CE approach includes the diacritic in the evaluation.
The presence or absence of the diacritic on the last letter mostly depends on grammatical rules.
Error rates without CE reflect the performance specifically on the core word, while error rates with CE represent the overall performance of the model \cite{madhfar2020effective}.

We compared our models with 9 models, namely, CBHG \cite{madhfar2020effective}, Sakhr\footnote{https://tashkeel.alsharekh.org}, Farasa \cite{darwish2016farasa}, D2 and D3 models \cite{alkhamissi-etal-2020-dd}, Alkhalil Tashkeel\footnote{https://tashkeel.alkhalilarabic.com}, Mishkal\footnote{https://tahadz.com/mishkal}, Multilevel Diacritizer \cite{abbad2021simple}, and Shakkala.
Moreover, we compared the performance of our models with the performance of two Large Language Models (LLMs) on the ATD task: GPT-4-turbo\footnote{https://chatgpt.com}, and Command R+\footnote{https://huggingface.co/CohereForAI/c4ai-command-r-plus}.
During evaluation, all preprocessing steps are applied to both the reference text and the output of all models to ensure fair comparisons.
In the case of long sentence diacritization, we follow a process of splitting the sentence into smaller segments based on punctuation.
Each small sentence is then diacritized individually, and finally, the segments are combined to reconstruct the original text with diacritics.

The results in Tables \ref{tab: catt_benchamrk_results} and \ref{tab:wikinews_results} show that our models achieved state-of-the-art performance in the ATD task, outperforming all 11 models.
Our ED model achieved the lowest DER and Word Error Rate (WER) on the CATT benchmark dataset with and without CE.
Moreover, both ED and EO models achieved low DER and WER scores on the WikiNews benchmark dataset compared to other ATD models.

The evaluation of GPT-4 on the WikiNews dataset shows a high level of performance.
However, when GPT-4 is evaluated on the CATT dataset, its performance appears to be comparatively normal.
This difference in performance can likely be attributed to the fact that GPT-4 was trained on web data predating December 2023.
Based on the observed significant performance gap between testing GPT-4 on CATT and WikiNews, we believe that GPT-4 was most likely trained on the WikiNews dataset.
It is worth noting that the WikiNews dataset was published in 2017 \cite{darwish2017arabic}, while the CATT dataset was created in March 2024.

\begin{table}[ht!]
\small
\begin{tabular}{lcccc}
\hline
\multicolumn{5}{c}{\multirow{2}{*}{\textbf{CATT Benchmark Dataset}}} \\
\multicolumn{5}{c}{} \\
\hline
\multicolumn{1}{l|}{\multirow{2}{*}{\textbf{Model}}} & \multicolumn{2}{c|}{\textbf{CE (\%)}} & \multicolumn{2}{c}{\textbf{No CE (\%)}} \\ \cline{2-5} 
\multicolumn{1}{c|}{} & \multicolumn{1}{c|}{\textbf{DER}} & \multicolumn{1}{c|}{\textbf{WER}} & \multicolumn{1}{c|}{\textbf{DER}} & \textbf{WER} \\ \hline
\multicolumn{1}{l|}{CBHG} & \multicolumn{1}{c|}{10.808} & \multicolumn{1}{c|}{42.680} & \multicolumn{1}{c|}{8.313} & 34.386 \\
\multicolumn{1}{l|}{Command R+} & \multicolumn{1}{c|}{13.169} & \multicolumn{1}{c|}{48.518} & \multicolumn{1}{c|}{11.329} & 44.158 \\
\multicolumn{1}{l|}{GPT-4} & \multicolumn{1}{c|}{9.515} & \multicolumn{1}{c|}{38.311} & \multicolumn{1}{c|}{8.113} & 33.505 \\
\multicolumn{1}{l|}{Sakhr} & \multicolumn{1}{c|}{13.841} & \multicolumn{1}{c|}{56.661} & \multicolumn{1}{c|}{11.125} & 47.993 \\
\multicolumn{1}{l|}{Farasa} & \multicolumn{1}{c|}{17.825} & \multicolumn{1}{c|}{65.783} & \multicolumn{1}{c|}{15.414} & 60.114 \\
\multicolumn{1}{l|}{D2} & \multicolumn{1}{c|}{13.310} & \multicolumn{1}{c|}{49.417} & \multicolumn{1}{c|}{10.036} & 38.391 \\
\multicolumn{1}{l|}{D3} & \multicolumn{1}{c|}{58.313} & \multicolumn{1}{c|}{98.710} & \multicolumn{1}{c|}{48.018} & 95.186 \\
\multicolumn{1}{l|}{Alkhalil} & \multicolumn{1}{c|}{14.232} & \multicolumn{1}{c|}{53.413} & \multicolumn{1}{c|}{11.568} & 45.777 \\
\multicolumn{1}{l|}{Mishkal} & \multicolumn{1}{c|}{16.482} & \multicolumn{1}{c|}{60.844} & \multicolumn{1}{c|}{10.796} & 40.215 \\
\multicolumn{1}{l|}{Multilevel} & \multicolumn{1}{c|}{16.503} & \multicolumn{1}{c|}{58.076} & \multicolumn{1}{c|}{13.434} & 50.147 \\
\multicolumn{1}{l|}{Shakkala} & \multicolumn{1}{c|}{13.494} & \multicolumn{1}{c|}{50.387} & \multicolumn{1}{c|}{10.386} & 40.643 \\
\hline
\multicolumn{1}{l|}{EO (Ours)} & \multicolumn{1}{c|}{8.762} & \multicolumn{1}{c|}{35.508} & \multicolumn{1}{c|}{7.088} & 29.714 \\
\multicolumn{1}{l|}{\textbf{ED (Ours)}} & \multicolumn{1}{c|}{\textbf{8.624}} & \multicolumn{1}{c|}{\textbf{34.191}} & \multicolumn{1}{c|}{\textbf{6.989}} & \textbf{28.477}
\end{tabular}
\caption{Benchmark results on CATT dataset.}
\label{tab: catt_benchamrk_results}
\end{table}

\begin{table}[ht!]
\small
\begin{tabular}{l|cc|cc}
\hline
\multicolumn{5}{c}{\multirow{2}{*}{\textbf{WikiNews Benchmark Dataset}}} \\
\multicolumn{5}{c}{} \\
\hline
\multirow{2}{*}{\textbf{Model}} & \multicolumn{2}{c|}{\textbf{CE (\%)}} & \multicolumn{2}{c}{\textbf{No CE (\%)}} \\ \cline{2-5} 
& \multicolumn{1}{c|}{\textbf{DER}} & \textbf{WER} & \multicolumn{1}{c|}{\textbf{DER}} & \textbf{WER} \\ \hline
CBHG & \multicolumn{1}{c|}{8.276} & 36.032 & \multicolumn{1}{c|}{5.448} & 21.528 \\
Command R+ & \multicolumn{1}{c|}{21.470} & 54.335 & \multicolumn{1}{c|}{17.755} & 49.611 \\
GPT-4 & \multicolumn{1}{c|}{\textbf{0.551*}} & \textbf{2.276*} & \multicolumn{1}{c|}{\textbf{0.326*}} & \textbf{1.024*} \\
Sakhr & \multicolumn{1}{c|}{7.843} & 38.413 & \multicolumn{1}{c|}{5.628} & 29.921 \\
Farasa & \multicolumn{1}{c|}{19.584} & 69.536 & \multicolumn{1}{c|}{17.752} & 66.601 \\
D2 & \multicolumn{1}{c|}{9.231} & 32.622 & \multicolumn{1}{c|}{6.164} & 21.744 \\
D3 & \multicolumn{1}{c|}{58.558} & 109.028 & \multicolumn{1}{c|}{48.234} & 99.408 \\
Alkhalil & \multicolumn{1}{c|}{14.912} & 46.793 & \multicolumn{1}{c|}{12.145} & 35.699 \\
Mishkal & \multicolumn{1}{c|}{15.246} & 54.187 & \multicolumn{1}{c|}{8.558} & 27.337 \\
Multilevel & \multicolumn{1}{c|}{12.431} & 45.054 & \multicolumn{1}{c|}{9.318} & 36.217 \\
Shakkala & \multicolumn{1}{c|}{9.978} & 37.241 & \multicolumn{1}{c|}{6.593} & 24.988 \\
\hline
EO (Ours) & \multicolumn{1}{c|}{\textbf{5.425}} & 22.132 & \multicolumn{1}{c|}{\textbf{3.105}} & 12.679 \\
\textbf{ED (Ours)} & \multicolumn{1}{c|}{5.963} & \textbf{20.060} & \multicolumn{1}{c|}{3.631} & \textbf{11.310}
\end{tabular}
\caption{Benchmark results on WikiNews dataset.}
\label{tab:wikinews_results}
\end{table}

\subsection{RQ1: Does the ATD model benefit from MLM pretraining?}

Tables \ref{tab:mlm_vs_from_scratch_catt_long_training} through \ref{tab:training_techniques_wikinews} detail our experiments' results on the CATT and WikiNews datasets.
The experiments clearly show the advantages of MLM pretraining since it consistently boosted performance across all tested models, regardless of the number of training steps, the CE conditions, or the used benchmark dataset.
Table \ref{tab:mlm_vs_from_scratch_catt_long_training} indicates that initializing the encoder part of the ED model with pretrained MLM weights boosted the performance by a relative ratio of 12.28\% when evaluated on the CATT dataset. 
Similarly, Table \ref{tab:mlm_vs_from_scratch_wikinews_long_training} shows that initializing the EO model with pretrained MLM weights improved the performance by a relative ratio of 15.94\% when evaluated on the WikiNews dataset. 
Our results indicate that weight initialization from pretrained MLM weights can boost the performance of both ED and EO models compared to random initialization.

\subsection{RQ2: Does training ATD model for more
iterations help?}

By analyzing the results presented in Tables \ref{tab:mlm_vs_from_scratch_catt_long_training} and \ref{tab:mlm_vs_from_scratch_wikinews_long_training}, and Tables \ref{tab:mlm_vs_from_scratch_catt_short_training} and \ref{tab:mlm_vs_from_scratch_wikinews_short_training}, we can see that the training for more iterations consistently enhances model performance across all metrics, CE conditions, and datasets.
It is notable that for few training iterations, the EO model outperforms the ED model on the CATT benchmark dataset as shown in Tables \ref{tab:mlm_vs_from_scratch_catt_long_training} and \ref{tab:mlm_vs_from_scratch_catt_short_training}.
In the other hand, when we trained both models for more iterations, the ED model surpasses the EO model in all performance metrics on the same dataset.
Moreover, Tables \ref{tab:mlm_vs_from_scratch_wikinews_long_training} and \ref{tab:mlm_vs_from_scratch_wikinews_short_training} show the performance of both models on the WikiNews dataset.
They demonstrate that with fewer training steps, the EO model performs better in terms of DER and WER under both CE and No CE conditions. However, only the DER for the EO model was better than the DER of the ED model as we trained them for more iterations.
The performance difference between EO and ED could be attributed to the variation in model architecture.
Specifically, the EO model had all 6 layers initialized using pretrained MLM weights, while in ED, only the 3 layers of the encoder part were initialized with pretrained MLM weights.
Generally, our experiments show that training for more iterations can improve the ATD model performance.

\begin{table}[ht!]
\small
\begin{tabular}{lcccc}
\hline
\multicolumn{5}{c}{\multirow{2}{*}{\textbf{CATT Benchmark Dataset}}} \\
\multicolumn{5}{c}{} \\ \hline
\multicolumn{1}{l|}{\multirow{2}{*}{\textbf{Model}}} & \multicolumn{2}{c|}{\textbf{CE (\%)}} & \multicolumn{2}{c}{\textbf{No CE (\%)}} \\ \cline{2-5} 
\multicolumn{1}{l|}{} & \multicolumn{1}{c|}{\textbf{DER}} & \multicolumn{1}{c|}{\textbf{WER}} & \multicolumn{1}{c|}{\textbf{DER}} & \textbf{WER} \\ \hline 
\multicolumn{1}{l|}{EO – From Scratch} & \multicolumn{1}{c|}{9.613} & \multicolumn{1}{c|}{38.685} & \multicolumn{1}{c|}{7.631} & 31.610 \\
\multicolumn{1}{l|}{EO – MLM} & \multicolumn{1}{c|}{9.260} & \multicolumn{1}{c|}{37.492} & \multicolumn{1}{c|}{7.411} & 30.969 \\
\hline
\multicolumn{1}{l|}{ED – From Scratch} & \multicolumn{1}{c|}{10.359} & \multicolumn{1}{c|}{39.753} & \multicolumn{1}{c|}{8.003} & 31.654 \\
\multicolumn{1}{l|}{ED – MLM} & \multicolumn{1}{c|}{\textbf{9.087}} & \multicolumn{1}{c|}{\textbf{35.757}} & \multicolumn{1}{c|}{\textbf{7.272}} & \textbf{29.483}
\end{tabular}
\caption{The impact of MLM pretraining vs. training from scratch, after training for \textit{more} iterations when evaluated on CATT benchmark dataset.}
\label{tab:mlm_vs_from_scratch_catt_long_training}
\end{table}

\begin{table}[ht!]
\small
\begin{tabular}{lcccc}
\hline
\multicolumn{5}{c}{\multirow{2}{*}{\textbf{WikiNews Benchmark Dataset}}} \\
\multicolumn{5}{c}{} \\ \hline
\multicolumn{1}{l|}{\multirow{2}{*}{\textbf{Model}}} & \multicolumn{2}{c|}{\textbf{CE (\%)}} & \multicolumn{2}{c}{\textbf{No CE (\%)}} \\ \cline{2-5}
\multicolumn{1}{l|}{} & \multicolumn{1}{c|}{\textbf{DER}} & \multicolumn{1}{c|}{\textbf{WER}} & \multicolumn{1}{c|}{\textbf{DER}} & \textbf{WER} \\ \cline{1-5} 
\multicolumn{1}{l|}{EO – From Scratch} & \multicolumn{1}{c|}{7.009} & \multicolumn{1}{c|}{25.857} & \multicolumn{1}{c|}{4.527} & 16.761 \\
\multicolumn{1}{l|}{EO – MLM} & \multicolumn{1}{c|}{\textbf{5.892}} & \multicolumn{1}{c|}{23.785} & \multicolumn{1}{c|}{\textbf{3.469}} & 14.116 \\
\hline
\multicolumn{1}{l|}{ED – From Scratch} & \multicolumn{1}{c|}{7.271} & \multicolumn{1}{c|}{24.870} & \multicolumn{1}{c|}{4.464} & 14.202 \\
\multicolumn{1}{l|}{ED – MLM} & \multicolumn{1}{c|}{6.376} & \multicolumn{1}{c|}{\textbf{21.954}} & \multicolumn{1}{c|}{4.001} & \textbf{12.926}
\end{tabular}
\caption{The impact of MLM pretraining vs. training from scratch, after training for \textit{more} iterations when evaluated on WikiNews benchmark dataset.}
\label{tab:mlm_vs_from_scratch_wikinews_long_training}
\end{table}

\begin{table}[ht!]
\small
\begin{tabular}{lcccc}
\hline
\multicolumn{5}{c}{\multirow{2}{*}{\textbf{CATT Benchmark Dataset}}} \\
\multicolumn{5}{c}{} \\ \hline
\multicolumn{1}{l|}{\multirow{2}{*}{\textbf{Model}}} & \multicolumn{2}{c|}{\textbf{CE (\%)}} & \multicolumn{2}{c}{\textbf{No CE (\%)}} \\ \cline{2-5} 
\multicolumn{1}{l|}{} & \multicolumn{1}{c|}{\textbf{DER}} & \multicolumn{1}{c|}{\textbf{WER}} & \multicolumn{1}{c|}{\textbf{DER}} & \textbf{WER} \\ 
\hline
\multicolumn{1}{l|}{EO – From Scratch} & \multicolumn{1}{c|}{11.199} & \multicolumn{1}{c|}{43.294} & \multicolumn{1}{c|}{8.725} & 34.751 \\
\multicolumn{1}{l|}{EO – MLM} & \multicolumn{1}{c|}{\textbf{10.037}} & \multicolumn{1}{c|}{\textbf{39.904}} & \multicolumn{1}{c|}{\textbf{7.919}} & \textbf{32.598} \\
\hline
\multicolumn{1}{l|}{ED – From Scratch} & \multicolumn{1}{c|}{13.208} & \multicolumn{1}{c|}{48.171} & \multicolumn{1}{c|}{9.913} & 36.985 \\
\multicolumn{1}{l|}{ED – MLM} & \multicolumn{1}{c|}{11.345} & \multicolumn{1}{c|}{42.841} & \multicolumn{1}{c|}{8.805} & 34.066
\end{tabular}
\caption{The impact of MLM pretraining vs. training from scratch, after training for \textit{fewer} iterations when evaluated on CATT benchmark dataset.}
\label{tab:mlm_vs_from_scratch_catt_short_training}
\end{table}

\begin{table}[ht!]
\small
\begin{tabular}{lcccc}
\hline
\multicolumn{5}{c}{\multirow{2}{*}{\textbf{WikiNews Benchmark Dataset}}} \\
\multicolumn{5}{c}{} \\ \hline
\multicolumn{1}{l|}{\multirow{2}{*}{\textbf{Model}}} & \multicolumn{2}{c|}{\textbf{CE (\%)}} & \multicolumn{2}{c}{\textbf{No CE (\%)}} \\ \cline{2-5} 
\multicolumn{1}{l|}{} & \multicolumn{1}{c|}{\textbf{DER}} & \multicolumn{1}{c|}{\textbf{WER}} & \multicolumn{1}{c|}{\textbf{DER}} & \textbf{WER} \\ \hline
\multicolumn{1}{l|}{EO – From Scratch} & \multicolumn{1}{c|}{8.226} & \multicolumn{1}{c|}{29.958} & \multicolumn{1}{c|}{5.232} & 18.716 \\
\multicolumn{1}{l|}{EO – MLM} & \multicolumn{1}{c|}{\textbf{6.915}} & \multicolumn{1}{c|}{\textbf{26.221}} & \multicolumn{1}{c|}{\textbf{4.263}} & \textbf{16.120} \\
\hline
\multicolumn{1}{l|}{ED – From Scratch} & \multicolumn{1}{c|}{10.518} & \multicolumn{1}{c|}{34.805} & \multicolumn{1}{c|}{6.999} & 22.478 \\
\multicolumn{1}{l|}{ED – MLM} & \multicolumn{1}{c|}{8.681} & \multicolumn{1}{c|}{29.224} & \multicolumn{1}{c|}{5.741} & 18.617
\end{tabular}
\caption{The impact of MLM pretraining vs. training from scratch, after training for \textit{fewer} iterations when evaluated on WikiNews benchmark dataset.}
\label{tab:mlm_vs_from_scratch_wikinews_short_training}
\end{table}

\subsection{RQ3: Is the NS approach effective
in ATD models?}

We tested the impact of the NS approach on both the EO and ED models as shown in Tables \ref{tab:training_techniques_catt} and \ref{tab:training_techniques_wikinews}.
Our results show that NS approach further improved both ED and EO models, showing a considerable reduction in both DER and WER.
After evaluating on the CATT dataset, the ED model achieved the best overall performance.
However, the EO model outperformed the ED model specifically in DER under all CE conditions when evaluated on the WikiNews dataset.

\begin{table}
\small
\centering
\begin{tabular}{l|cccc}
\hline
\multicolumn{5}{c}{\multirow{2}{*}{\textbf{CATT Benchmark Dataset}}} \\
\multicolumn{5}{c}{} \\ \hline
\multirow{2}{*}{\textbf{Model}} & \multicolumn{2}{c|}{\textbf{CE (\%)}} & \multicolumn{2}{c}{\textbf{No CE (\%)}} \\ \cline{2-5} 
{} & \multicolumn{1}{c|}{\textbf{DER}} & \multicolumn{1}{c|}{\textbf{WER}} & \multicolumn{1}{c|}{\textbf{DER}} & \textbf{WER} \\ \hline
EO + Long Training & \multicolumn{1}{|c|}{9.613} & \multicolumn{1}{c|}{38.685} & \multicolumn{1}{c|}{7.631} & 31.610 \\
\hspace{7mm} + MLM & \multicolumn{1}{c|}{9.260} & \multicolumn{1}{c|}{37.492} & \multicolumn{1}{c|}{7.411} & 30.969 \\
\hspace{14mm}+ NS & \multicolumn{1}{c|}{8.762} & \multicolumn{1}{c|}{35.508} & \multicolumn{1}{c|}{7.088} & 29.714 \\ \hline
ED + Long Training & \multicolumn{1}{c|}{10.359} & \multicolumn{1}{c|}{39.753} & \multicolumn{1}{c|}{8.003} & 31.654 \\
\hspace{7mm} + MLM & \multicolumn{1}{c|}{9.087} & \multicolumn{1}{c|}{35.757} & \multicolumn{1}{c|}{7.272} & 29.483 \\
\hspace{14mm}+ NS & \multicolumn{1}{c|}{\textbf{8.624}} & \multicolumn{1}{c|}{\textbf{34.191}} & \multicolumn{1}{c|}{\textbf{6.989}} & \textbf{28.477}
\end{tabular}
\caption{Performance comparison of our training techniques on the CATT benchmark dataset.}
\label{tab:training_techniques_catt}
\end{table}

\begin{table}
\small
\centering
\begin{tabular}{l|cccc}
\hline
\multicolumn{5}{c}{\multirow{2}{*}{\textbf{WikiNews Benchmark Dataset}}} \\
\multicolumn{5}{c}{} \\ \hline
{\multirow{2}{*}{\textbf{Model}}} & \multicolumn{2}{c|}{\textbf{CE (\%)}} & \multicolumn{2}{c}{\textbf{No CE (\%)}} \\ \cline{2-5} 
{} & \multicolumn{1}{c|}{\textbf{DER}} & \multicolumn{1}{c|}{\textbf{WER}} & \multicolumn{1}{c|}{\textbf{DER}} & \textbf{WER} \\ \hline
EO + Long Training & \multicolumn{1}{|c|}{7.009} & \multicolumn{1}{c|}{25.857} & \multicolumn{1}{c|}{4.527} & 16.761 \\
\multicolumn{1}{l|}{\hspace{7mm} + MLM} & \multicolumn{1}{c|}{5.892} & \multicolumn{1}{c|}{23.785} & \multicolumn{1}{c|}{3.469} & 14.116 \\
\multicolumn{1}{l|}{\hspace{14mm} + NS} & \multicolumn{1}{c|}{\textbf{5.425}} & \multicolumn{1}{c|}{22.132} & \multicolumn{1}{c|}{\textbf{3.105}} & 12.679 \\ \hline
\multicolumn{1}{l|}{ED + Long Training} & \multicolumn{1}{c|}{7.271} & \multicolumn{1}{c|}{24.870} & \multicolumn{1}{c|}{4.464} & 14.202 \\
\multicolumn{1}{l|}{\hspace{7mm} + MLM} & \multicolumn{1}{c|}{6.376} & \multicolumn{1}{c|}{21.954} & \multicolumn{1}{c|}{4.001} & 12.926 \\
\multicolumn{1}{l|}{\hspace{14mm} + NS} & \multicolumn{1}{c|}{5.963} & \multicolumn{1}{c|}{\textbf{20.060}} & \multicolumn{1}{c|}{3.631} & \textbf{11.310}
\end{tabular}
\caption{Performance comparison of our training techniques on the WikiNews benchmark dataset.}
\label{tab:training_techniques_wikinews}
\end{table}

\section{Conclusion}
This paper proposed a new approach to training ATD models.
The proposed approach was to initialize ATD models' parameters from a pretrained character-based BERT model, then training the models for longer iterations.
After that, we used the NS approach to further improve the performance of our models.
We evaluated our approach by comparing it to 11 commercial and open-source models using two benchmark datasets: WikiNews and CATT.
Our results show that our models outperformed all other models in both DER and WER.
We open-source our CATT models and dataset for the research community to advance research in this area.

\section*{Limitations}

Although this research advances the progress in the ATD task, it has some limitations.
These limitations include:
\begin{itemize}
    \item \textbf{Specific input assumption}: Our model is designed to work only with Arabic text. It does not handle numbers or special characters often found in real-world data.
    Filtering out these unwanted characters and numbers may alter the sentence structure, potentially resulting in incorrect ground truth diacritics.
    For example, consider the sentence \<اشتريت 3 كتب> (translation: I bought 3 books).
    If the number \<3> is removed, the sentence becomes \<اشتريت كتب>, which results in an incorrect grammatical structure (correct structure: \<اشتريت كتبا>) that may also lead to incorrect diacritization.
    A correct normalization should replace the numeral with an equivalent Arabic word.
    Meaning, the sentence should be transformed to \<اشتريت ثلاثة كتب>, where \<ثلاثة> represents the number 3 in Arabic.
    Therefore, it is suggested to have a normalization layer before the diacritizing text in a sequential pipeline.
    
    \item \textbf{No handling for partially diacritized input}: The models are not conditioned to process partially diacritized text, as we filter those diacritics out before the text is fed into the model.
\end{itemize}

\bibliography{emnlp2023}
\bibliographystyle{acl_natbib}




\end{document}